\documentclass[11pt]{article}

\usepackage[final]{acl}

\usepackage{times}
\usepackage{latexsym}
\usepackage{booktabs}
\usepackage{multirow}
\usepackage[T1]{fontenc}

\usepackage[utf8]{inputenc}

\usepackage{microtype}

\usepackage{inconsolata}

\usepackage{graphicx}

\usepackage[table]{xcolor}
\usepackage{array}

\newcolumntype{H}{>{\columncolor{orange!15}}c}

\usepackage{array}
\usepackage{enumitem}
\usepackage[T1]{fontenc}    
\usepackage{hyperref}       
\usepackage{url}            
\usepackage{booktabs}       
\usepackage{amsfonts}       
\usepackage{nicefrac}       
\usepackage{microtype}      
\usepackage{xcolor}         
\usepackage{amsmath}
\usepackage{amssymb}
\usepackage{graphicx}
\usepackage{caption} 
\usepackage{booktabs}
\usepackage{wrapfig} 
\usepackage{subcaption}
\usepackage{tcolorbox}

\newcommand\blfootnote[1]{%
  \begingroup
  \renewcommand\thefootnote{}\footnote{#1}%
  \addtocounter{footnote}{-1}%
  \endgroup
}

\newcommand{\researchquestionbox}[2]{
    {
    \vspace{-0mm}
    \begin{tcolorbox}[
        colback=#2!3,
        colframe=#2!50!black,
        boxrule=1.4pt,
        right=1mm,
        left=1mm,
        bottom=2mm,
        top=2mm
    ]
        \fontsize{10.5pt}{11pt}\selectfont
        \vspace{0mm}
        #1
        
        \vspace{0mm}
    \end{tcolorbox}
    }
    \vspace{-0mm}
}

\newtcolorbox[auto counter]{promptbox}[2][]{
  colback=gray!5,
  colframe=gray!50!black,
  fonttitle=\bfseries,
  title={Prompt \thetcbcounter: #2},
  boxrule=0.5pt,
  arc=2pt,
  left=6pt, right=6pt, top=4pt, bottom=4pt,
  #1
}

\title{Do LLMs Exhibit Coherent Knowledge Structures in Mathematical Reasoning? A Perspective from Knowledge Space Theory}

\author{
Peng Cui$^{*1}$, Heejin Do$^{*2}$,  Mrinmaya Sachan$^{1}$ \\
ETH Zürich Department of Computer Science$^{1}$, ETH AI Center$^{2}$ \\ 
\texttt{
\{peng.cui, mrinmaya.sachan\}@inf.ethz.ch$^1$} \texttt{heejindo@ai.ethz.ch}$^2$ \\
}

\begin{document}
\maketitle

\begin{abstract}
Human knowledge is inherently structured and interdependent: mastery of a concept requires prior mastery of its prerequisites, a principle formalized by Knowledge Space Theory (KST). While LLMs achieve strong performance on complex reasoning tasks, it remains unclear whether they exhibit coherent, human-like knowledge structure. We introduce a KST-grounded framework for evaluating LLM knowledge structure in mathematical reasoning, using it as a normative framework to analyze whether LLM behavior adheres to principled knowledge dependencies. Evaluating eight open- and closed-source LLMs against real human learners, we find that (1) LLMs do not adhere to human knowledge structure—they frequently violate knowledge dependencies and fail to leverage related knowledge provided in context to improve performance on dependent questions; (2) LLMs do not share a consistent knowledge structure among themselves, as reflected by low overlap in their knowledge distributions. 
Furthermore, these structural deficiencies remain largely invisible to accuracy-based and LLM-as-judge evaluations. 
Together, our results provide behavioral evidence that current LLMs knowledge does not follow a human-like structure. \blfootnote{*: Equal contribution}
\newline
\newline
    \vspace{0.2em}
    \raisebox{-1mm}{
    \includegraphics[width=1.05em,height=1.05em]{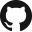}}
    \hspace{.5em}\parbox{\dimexpr\linewidth-2\fboxsep-2\fboxrule}
    {\small \href{https://github.com/pengcuix/LLM-KST}{https://github.com/pengcuix/LLM-KST}}
\end{abstract}

\section{Introduction}

\begin{figure}[t]
\centering
\includegraphics[width=\columnwidth]{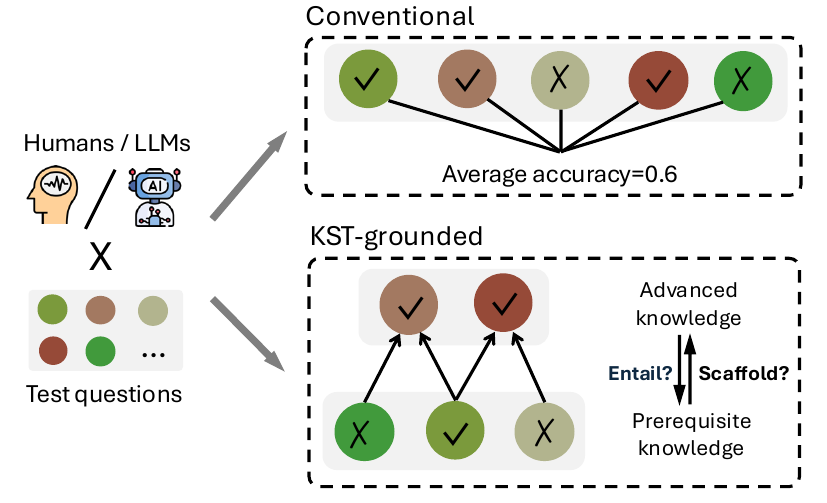}
\caption{Conventional evaluation measures average accuracy, treating knowledge as a flat unstructured collection. 
We propose a KST-based framework that models structured dependencies in knowledge, examining whether LLM exhibits coherent knowledge behavior.} \label{fig:intro}

\end{figure}

Recent advances in large language models (LLMs) have led to remarkable performance on a wide range of reasoning benchmarks. 
Yet high accuracy alone does not imply genuine understanding. 
A growing body of evidence shows that models can arrive at correct answers through flawed, short-cut-based, or unfaithful reasoning processes \cite{lanham2023measuring, turpin2023language}. In response, recent work has shifted from evaluating final answers to evaluating reasoning trajectories themselves, introducing process-based benchmarks \cite{lightman2023let,uesato2022solving,xia2025evaluating,do2025defines}.
However, these evaluations remain inherently \emph{local}: they assess each problem in isolation and cannot reveal whether a model's success and failure patterns are globally consistent with the structure of the knowledge being tested.

In contrast, established theories of human learning emphasize that knowledge is inherently \emph{structured}. 
Knowledge Space Theory (KST)~\cite{doignon2012knowledge} formalizes this principle by modeling a knowledge domain as a set of latent concepts connected through prerequisite 
dependencies. 
Under this framework, mastery of a concept presupposes mastery of its prerequisites — for instance, solving quadratic equations requires prior mastery of linear equations and algebraic manipulation. 
These dependency relations constrain what constitutes a valid knowledge state (i.e., which concepts a learner has mastered) and define coherent learning pathways through the domain.

In this work, we adopt KST as a normative framework for evaluating the behavioral consistency of LLMs. 
We ground our study in mathematics, where prerequisite relations are well-defined and extensively documented through expert-curated educational standards. 
As illustrated in Figure~\ref{fig:intro}, rather than the conventional approach of treating knowledge as a flat collection of independent items and measuring individual or aggregate accuracy, our framework analyzes whether the knowledge structures underlying observed LLM response patterns adhere to principled dependencies. 
Specifically, we aim to answer two research questions: 
\begin{itemize}[topsep=4pt, itemsep=2pt, leftmargin=20pt]
\item \textbf{RQ1}: Do LLMs adhere to human knowledge dependencies? 
\item \textbf{RQ2}: If not, do LLMs share a consistent and coherent knowledge structure among 
themselves?
\end{itemize}

We evaluate a broad range of open- and closed-source LLMs alongside human learners on a dataset with real student response records. Our results show that: 
(1) despite a moderate accuracy of 79.6\%, human learners satisfy prerequisite dependencies for the majority (72.7\%) of their correct answers. LLMs also fail to leverage prerequisite knowledge to scaffold dependent questions, further suggesting a lack of human-like knowledge structure. 
(2) Among human learners, stronger learners' knowledge states consistently subsume those of weaker ones, reflecting ordered knowledge growth. LLMs, in contrast, show significantly lower subsumption across performance levels, suggesting that LLMs do not share a consistent and coherent knowledge structure, and that their knowledge acquisition is more likely flat than structured.

In summary, our contributions are as follows:
\begin{itemize}[topsep=4pt, itemsep=2pt, leftmargin=20pt]
\item We propose a novel KST-grounded analytical framework for evaluating the coherence of LLM knowledge, providing a principled complement to accuracy-based evaluation. 
\item Our analysis reveals systematic incoherence in LLM mathematical knowledge structure, offering a new perspective and empirical evidence that current LLMs may not engage in genuine formal reasoning.
\item We construct a dataset with concept and dependency annotations, providing a resource for future research on LLM knowledge structure. 
\end{itemize}

\section{Related Work}

\paragraph{Knowledge Space Theory (KST)} is a theoretical framework for modeling the structure of knowledge and learning introduced by \citet{doignon1985spaces, doignon2012knowledge} in the 1980s. 
The central idea of KST is that learners’ knowledge is not an arbitrary collection of isolated facts, but rather forms a structured space constrained by prerequisite relations among concepts or skills. 
In this framework, each learner is associated with a knowledge state representing the subset of problems or concepts they have mastered, while the set of all feasible states forms a knowledge space. 
KST further models learning as transitions between knowledge states, thereby providing a principled representation of hierarchical and cumulative learning processes.

Subsequent work extended KST from the deterministic framework to probabilistic models to account for response noise \cite{falmagne1988class, de2024reliability}, introducing parameters such as lucky guesses and careless errors to model the stochastic nature of real-world assessment data, enabling KST to be applied to large-scale empirical settings.
Over the past decades, KST has been widely applied in educational assessment \cite{falmagne2013knowledge}, intelligent tutoring systems \cite{nkambou2010advances}, and adaptive learning environments \cite{falmagne2010learning}.
Early systems such as ALEKS demonstrated the practical utility of KST for personalized assessment and curriculum sequencing \cite{cosyn2021practical, cui-sachan-2023-adaptive}.

\paragraph{LLM Reasoning Evaluation} 

Recent efforts to evaluate the capabilities of LLMs have shifted from simple answer-correctness metrics to the scrutiny of reasoning trajectories that lead to the answer. Driven by the observation that correct final answers can emerge from flawed or unfaithful reasoning chains \cite{lanham2023measuring, turpin2023language}, the focus has moved toward assessing the validity of intermediate steps through process-based reward models (PRMs) \cite{lightman2023let, uesato2022solving} and targeted reasoning benchmarks \cite{cobbe2021training, NEURIPS2024_d81cb1f4}. Further attempts have dissected reasoning quality into multiple dimensions \cite{xia2025evaluating, do2025defines}, providing granular signals for model improvement. However, these trajectory-based evaluations remain inherently limited by their focus on local, isolated problem-solving; as such, they fail to capture whether a model's behavior is consistent with the latent hierarchical structure of knowledge. In human cognition, mastery is not a collection of isolated successes but a consistent epistemic state where complex concepts are built upon foundational prerequisites \cite{doignon2012knowledge}. By shifting the evaluative focus from superficial texts to structural dependencies, this work introduces a principled framework to detect these hidden inconsistencies, offering a more global and robust measure of model reliability.

\begin{figure*}[t]
\centering
\includegraphics[width=0.97\textwidth]{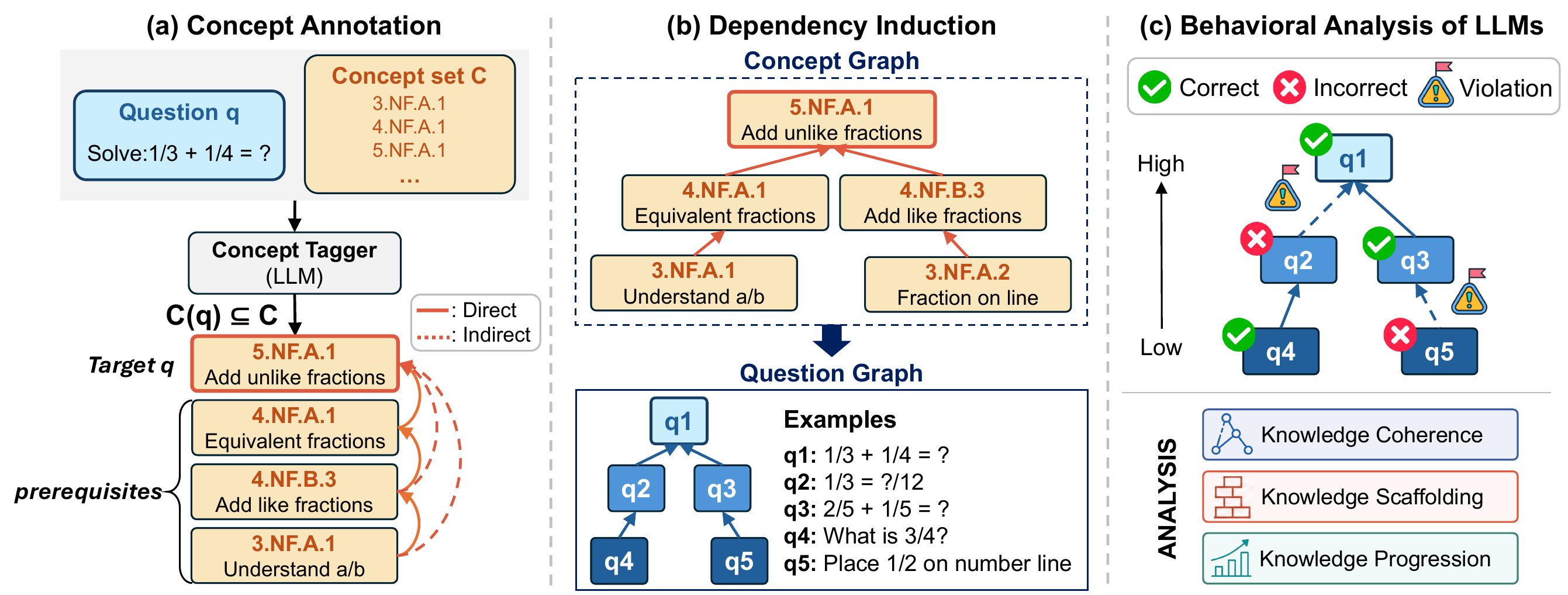}
\caption{\textbf{(a) Concept Annotation}: For each question, we prompt an advanced LLM to identify the relevant mathematical concepts from an 
expert-curated concept map.
\textbf{(b) Dependency Induction}: We use the question-concept associations together with expert-defined concept dependencies to infer prerequisite relations between questions.
\textbf{(c) Behavioral Analysis of LLMs}: We assess whether the knowledge structure and dynamics of LLMs are coherent with respect to the constructed knowledge space through a list of normative behaviors.}\label{fig1}

\end{figure*}

\section{Framework}
\paragraph{Overview.} 
In this section, we propose a normative framework grounded in Knowledge Space Theory (KST) to analyze whether the knowledge structure of LLMs is coherent. 
We begin by introducing the foundational definitions and concepts of KST (\S \ref{framework:background}), followed by a description of how we adapt this 
framework to LLMs under the mathematical reasoning setting (\S \ref{framework:construct}). 
Building on this, we define a list of normative behaviors characterizing the ideal knowledge state and dynamics in LLMs, along with metrics to quantitatively measure the degree to which LLMs conform to these behaviors (\S \ref{framework:behavior}). 
Together, these provide a principled basis for identifying incoherent aspects of LLM knowledge structure that would otherwise remain obscured by standard accuracy-based evaluation.
See Figure \ref{fig1} for an overview.

\subsection{Background of KST} \label{framework:background}

KST is a mathematical framework for modeling the structure of human knowledge.
It represents the domain of knowledge as a finite set of \textit{knowledge concepts} $C = \{c_1, \ldots, c_m\}$, where each concept is a fundamental unit of knowledge, serving as a building block for understanding and reasoning within a domain, for example, \textit{quadratic equations} in math. 
A central assumption of KST is that these concepts are not independent, i.e., mastery of some concepts is required before others can be acquired. 
This is formalized as a \textit{prerequisite relation}, a partial order $\preceq$ on $C$, where $c_i \preceq c_j$ denotes that $c_i$ is a prerequisite of $c_j$. 

A \textit{knowledge state} $K \subseteq C$ is the set of concepts a learner has mastered at a given point. 
Note that a valid knowledge state must be \textit{closed} under prerequisites: if $c_j \in K$ and $c_i \preceq c_j$, then $c_i\in K$. 
This closure property ensures that a learner cannot have mastered a concept without having mastered its prerequisites.
The collection of all valid knowledge states forms a \textit{knowledge space} $\mathcal{K}$.

In this work, we focus on mathematical reasoning, where knowledge is highly structured and interdependent, making it an ideal testbed for our framework. 
Many expert-defined mathematical standards with explicit prerequisite dependencies exist, such as the Common Core State Standards \cite{national2010common}. 
We use the New York State Mathematics Learning Standards\footnote{\href{https://www.nysed.gov/sites/default/files/programs/curriculum-instruction/nys-next-generation-mathematics-p-12-standards.pdf}{NYS Mathematics Learning Standards}} as our $C$, along with their defined dependency relations among concepts.

\subsection{Knowledge Space Construction} \label{framework:construct}

Since a learner’s mastery of knowledge concepts is difficult to estimate, we operationalize KST at the question level, where performance on each question can be directly observed and evaluated.
Let $Q = \{q_1, q_2, \ldots, q_n\}$ be a set of mathematical questions, where each question $q \in Q$ is associated with one or more concepts in $C$, denoted as $C(q)$. 
However, the dependencies among questions $Q$ are unknown. 
Therefore, we first infer question dependencies from their associated concepts.

\paragraph{LLM-based concept annotation} Specifically, we first prompt an LLM to identify the relevant concepts from $C$ for each question. 
Prior work has demonstrated that advanced LLMs can accurately identify 
required skills (equivalent to concepts) from question text, particularly in mathematical domains \cite{didolkar2024metacognitive, li2024automate,shah2024ai}. 
Unlike prior approaches, which rely on free-form concept generation, we prompt the LLM with the question together with a predefined concept list — the NYS standards — and instruct it to select relevant concepts only from this list. 
The NYS standards are organized hierarchically into 63 domains, 148 clusters, and 480 concepts, where each domain is a group of related clusters, and each cluster is a group of concepts. 
Supplying all 480 concepts and their descriptions in a single prompt would result in an unnecessarily long context, which increases cost and makes accurate selection more difficult. 
We therefore adopt a two-stage prompting procedure: the LLM first selects the relevant clusters, and is then prompted to choose the final concepts from those belonging to the selected clusters.
Our prompt template for concept annotation is in Prompt \ref{prompt:skill_extraction}.

\paragraph{Deriving question dependencies}
Grounded on concept-level dependencies, we define the prerequisite relation between questions as follows: question $q_i$ is a prerequisite of question $q_j$ if and only if
\begin{equation}
    \forall\, c_a \in C(q_i),\; \exists\, c_b \in C(q_j) 
    \text{ such that } c_a \preceq c_b.
\end{equation}
We intentionally adopt this strict definition, rather than weaker alternatives such as requiring only a single concept pair to satisfy the prerequisite relation. 
This ensures that the induced question-level dependencies are precise and conservative, reducing the risk of spurious prerequisites that could confound our analysis.

For each question $q$, we denote its prerequisite questions as ${\rm Pre}(q)$, which is a subset of $Q$.
For a learner $l$ (either an LLM or a human learner), we define their knowledge state $K_l$ as a subset of questions $Q$ that they answer correctly.

\subsection{Behavioral Analysis of LLMs via KST}\label{framework:behavior}
In this section, we use the KST framework to analyze whether LLMs follow a coherent knowledge structure. Specifically, we examine (RQ1) whether LLMs adhere to human knowledge dependencies, and (RQ2) if not, whether LLMs share a consistent knowledge structure among themselves. We investigate these two questions through three \textbf{normative behaviors (NBs)} that a reliable reasoning model should exhibit. 

\subsubsection{Do LLMs Adhere to Human Knowledge Dependencies? (RQ1)}
\paragraph{Should human knowledge structure apply to LLMs?}
Although the prerequisite dependencies among concepts in $C$ are derived from human knowledge standards, we argue that these dependencies should, in principle, apply to any reliable reasoning system. 
Mathematical reasoning follows strict logical procedures that are intrinsic to mathematics itself, not artifacts of human cognition. 
For example, an LLM that fails at $3 + 3$ (\texttt{Addition}) yet succeeds at $3 \times 3$ (\texttt{Multiplication}) is unlikely to arrive at the latter through genuine reasoning, and instead may rely on memorization or superficial pattern matching.
Therefore, LLMs' conformity to knowledge dependencies could serve as a signal of genuine logical reasoning.

\researchquestionbox{\textbf{NB1}: Mastery of a question $q$ should entail mastery of its prerequisite questions ${\rm Pre}(q)$.}{blue}
\noindent \textbf{NB1} is grounded in the foundational axiom of KST:  a valid knowledge state must be closed under prerequisite relations, that is, if a learner has mastered an item, they must have also mastered all of its prerequisites.
We measure the extent to which an LLM follows NB1 using the Prerequisite Satisfaction Ratio (PSR). 
Given a learner's knowledge state $K_{l}$, the PSR for a single question $q$ is defined as:
\begin{equation}
{\rm PSR} (q, K_l) = \frac{|\mathrm{Pre}(q) \cap K_{l}|}{\left|\mathrm{Pre}(q)\right|}, \label{eq:psr_q}
\end{equation}
which is the proportion of $q$'s prerequisite questions (${\rm Pre}(q)$) that the learner has also answered correctly. Note that PSR is computed only on correctly answered questions. 

Under NB1, this metric should be $\approx 1$ for all correctly answered questions.
In practice, however, exceptions can occur. 
For human learners, this may be attributed to a lucky guess on the target question, a careless slip on one or more prerequisite questions, or a prerequisite that is substantially more difficult than the target itself. 
Nevertheless, the overall PSR is expected to remain high. 
For LLMs, however, such violations probably suggest that correct answers may not be grounded in the requisite knowledge structure, but rather obtained through some surface-level pattern matching.

We aggregate the PSR of all questions $q \in K_l$ in either a macro- or micro-level manner:
\begin{gather}
    {\rm PSR}_{\rm macro}(K_l) = \frac{1}{|K_l|} \sum_{q \in {K_l}} {\rm PSR}(q), \label{eq:psr_macro} \\
    {\rm PSR}_{\rm micro}(K_l) = \frac{\sum_{q \in K_l} |\mathrm{Pre}(q) \cap K_l|}{\sum_{q \in K_l} \left|\mathrm{Pre}(q)\right|}, \label{eq:psr_micro}  
\end{gather}
where the macro-level PSR simply averages per-question PSR scores, and the micro-level PSR is a weighted average that accounts for the number of prerequisites per question.

\researchquestionbox{\textbf{NB2}: For a question $q$ with concept $c$, knowledge of $c$ or $c$'s prerequisite concepts ${\rm Pre}(c)$ should improve a learner's performance on $q$.}{orange}
\noindent \textbf{NB2} captures the functional role of relevant knowledge in scaffolding the acquisition of dependent or related concepts — a principle central to both KST and constructivist theories of learning \cite{narayan2013constructivism}.  
If LLMs internalize knowledge units and their relationships similarly to humans, we would expect to observe a scaffolding effect in LLMs as well.

We operationalize this for LLMs with In-Context Learning (ICL), where we provide questions and solutions of (1) prerequisite concepts or (2) the same concept as in-context examples. 
Let $K_l^{+\text{pre}}$ denote the knowledge state of the model when, for each question $q \in Q$, relevant examples are provided in context. 
We define Scaffolding Gain (SG) of a model $l$ as:
\begin{gather}
    \text{SG}(K_l) = \frac{1}{|Q|} ({|K_l^{+\text{pre}}| - |K_l|}), \label{eq6}
\end{gather}
which reflects the change in accuracy when relevant knowledge is provided as in-context scaffolding.

\subsubsection{Do LLMs Share a Coherent Knowledge Structure? (RQ2)}
Although human knowledge structure should in principle apply to LLMs, it is also possible that LLMs develop their own knowledge organization that is shared and consistent across models. If so, we would expect to observe the following behavior:
\vspace{-3mm}
\researchquestionbox{\textbf{NB3}: The knowledge state of a more capable model should largely subsume that of a less capable model.}{green}
\noindent \textbf{NB3} reflects the cumulative and hierarchical nature of mathematical knowledge. 
Unlike factual knowledge, which can often be acquired independently and in a fragmented manner, mathematical concepts are connected through dense prerequisite dependencies: advanced concepts build upon foundational ones and therefore require mastery of prior knowledge. 
Therefore, if LLMs possess a coherent and consistent knowledge structure of their own — even if it differs from that of humans — the knowledge state of a less capable model should still be largely subsumed by that of a more capable one, since greater competence presupposes mastery of the same underlying foundations.
In practice, the extent of such subsumption is expected to increase with the density of prerequisite dependencies in the domain. In tightly interconnected knowledge structures, there are fewer independent pathways to mastery, making coherent and nested knowledge states more likely to emerge. 

We use the Knowledge Overlap Coefficient (${\rm KOC}$) to quantify the degree of subsumption, defined as:
\begin{gather}
    {\rm KOC}(K_{l_1}, K_{l_2}) = \frac{| K_{l_1} \cap K_{l_2} |}{{\rm Min} (|K_{l_1}|, |K_{l_2}|)}, \label{eq:koc}
\end{gather}
where ${\rm KOC} \in [0, 1]$; a value of 0 indicates no overlap, while a value of 1 indicates that the weaker model's knowledge is fully subsumed by that of the stronger model. 
Under NB3, we would ideally expect ${\rm KOC} \approx1$. Note that this metric focuses solely on the alignment of question-level performance distributions between models, without making any assumptions about knowledge concepts or their dependencies.

However, since all $K_l$ are defined over a shared question set $Q$, KOC is inflated by chance overlap — the expected KOC between two randomly drawn knowledge states equals the accuracy of the stronger learner $ \frac{ {\rm max} (|K_{l_1}|, |K_{l_2}|)}{|Q|}$. 
We therefore compute a normalized variant:
\begin{equation}
    \text{KOC}_{\text{norm}}(K_{l_1}, K_{l_2}) = 
    \frac{\text{KOC}(K_{l_1}, K_{l_2}) - p_{\max}}{1 - p_{\max}}, \label{eq:koc_norm}
\end{equation}

\noindent where $p_{\max} = \max(|K_1|, |K_2|) / |Q|$ is the accuracy of the stronger learner. Under this  normalization, a value of 0 indicates overlap consistent  with chance, values greater than 0 indicate systematic  subsumption beyond chance, and a value of 1 indicates  perfect subsumption.

\section{Experimental Setup}

\paragraph{Datasets}

We conduct experiments on the mathematical knowledge tracing dataset XES3G5M \cite{liu2023xes3g5m}\footnote{\href{https://github.com/ai4ed/XES3G5M}{https://github.com/ai4ed/XES3G5M}}, using its English translated version publicly available by \citet{seo2026behavior}\footnote{\href{https://github.com/sjin4861/BAIM}{https://github.com/sjin4861/BAIM}}. 
Unlike standard math benchmarks, this dataset is derived from real student problem-solving logs and includes ground-truth correctness labels, enabling direct comparison between model predictions and human performance in terms of knowledge structure\footnote{To the best of our knowledge, this is the only publicly available mathematics dataset that contains both real student response records and full question content.}.
The dataset is used in compliance with the MIT License and its intended use.
As our focus is on textual knowledge, we exclude samples containing images and remove duplicate instances, resulting in 3,103 fill-in-the-blank questions (FITB) and 1,015 multiple-choice questions (MCQ).
The statistics of the data are summarized in Table \ref{tab:data_stats}.

\begin{table}[h]
\small
\centering
\begin{tabular}{lll}
\specialrule{1.0pt}{0pt}{2pt}
\specialrule{0.4pt}{0pt}{2pt}
\multirow{2}{*}{\textbf{Interactions}} & \# Students                       & $18,066$  \\
                              & \# Questions                      &  $4,118$ \\
                              \midrule
\multirow{2}{*}{\textbf{Concepts}}     & \# Concepts                       & $480$  \\
                              & \# Concepts / question (avg.)     & $1.07$  \\ \midrule
\multirow{2}{*}{\textbf{Dependency}}   & \# Prerequisites / concept (avg.)  & $1.58$ \\
                              & \# Prerequisites / question (avg.) & $16.4$ \\ 
\specialrule{0.4pt}{2pt}{0pt}
\specialrule{1.0pt}{2pt}{0pt}
\end{tabular}
\caption{Statistics of the XES3G5M dataset and extracted concepts.}\label{tab:data_stats}
\end{table}

\paragraph{LLMs and inference setup}
We evaluate a broad set of large language models spanning multiple families and scales. This includes strong closed-source models \textsc{Claude Sonnet 4.6}, \textsc{GPT-4.1-mini}~\cite{openai2025gpt41} and open-source models including \textsc{Mistral-7B-Instruct-v0.3}~\cite{jiang2023mistral}, \textsc{Llama-3.1-8B-Instruct}, \textsc{Llama-3.1-70B-Instruct}~\citep{grattafiori2024llama3}, \textsc{Qwen2.5-7B-Instruct}, \textsc{Qwen2.5-32B-Instruct}~\citep{qwen2025qwen25}, and \textsc{Qwen3-Next-80B-A3B-Instruct}~\citep{qwen2025qwen3next}. 

For open-source models, we perform inference using vLLM~\citep{kwon_vllm} on NVIDIA GH200 GPUs with temperature $0.6$, top-$p=0.95$, and random seed $42$. Closed-source models (Claude Sonnet 4.6 and GPT-4.1-mini) are accessed through their official APIs with default chat templates. We use a context window of $16{,}384$ tokens for the no-context baseline and $32{,}768$ tokens for all scaffolding-based evaluations.
All experiments are conducted using the FuseAI framework~\citep{wan2024knowledge}\footnote{\href{https://github.com/fanqiwan/FuseAI}{https://github.com/fanqiwan/FuseAI}}, with its default chain-of-thought (CoT; \citet{wei2022chain}) prompting templates (Prompt \ref{prompt:baseline_fitb}) applied consistently across all models. 

\paragraph{Concept annotation}
The prompt template for concept annotation is in Prompt \ref{prompt:skill_extraction}.
We use \textsc{GPT-4.1-mini} for concept extraction with a temperature of $0.2$. 
Since LLM annotations may contain errors, we sampled 400 questions for manual verification in order to quantify the error rate and its potential impact on our results. 
Of these, 317 were confirmed to be correctly annotated, giving an accuracy of 79.3\%. We denote the fully automatically annotated dataset as XES$_{\rm full}$ and the verified subset as XES$_{\rm verified}$. We additionally verified annotation accuracy on the exemplar questions from the NYS standards; details are given in Appendix \ref{app:annotation_eval}.

\paragraph{Comparison against human learners}
Our evaluation measures the extent to which LLMs conform to or deviate from the three normative behaviors using the corresponding metrics proposed in 
Section \S \ref{framework:behavior}. 
As discussed, these behaviors represent idealized conditions, but exceptions can happen in practice. 
To provide a meaningful reference point, we compare the conformance of both humans and LLMs with respect to each normative behavior, allowing us to quantify the gap, or potential advantage, between LLMs and human minds as a reliable reasoning system.

\begin{table*}[ht]
\small
\centering
\setlength{\tabcolsep}{3pt}
\begin{tabular}{lc|cc|cccccH}
\specialrule{1.0pt}{0pt}{2pt}
\specialrule{0.4pt}{0pt}{2pt}
\multirow{2}{*}{\textbf{Model}} & \multicolumn{1}{l|}{\multirow{2}{*}{\textbf{Acc.}}}  & \multicolumn{2}{c|}{\textbf{Avg. PSR}}  & \multicolumn{6}{c}{\textbf{Question-level PSR distribution}}                                                                  \\
                                & \multicolumn{1}{l|}{}                                & \textbf{Micro}       & \textbf{Macro}        & $\mathbf{[0, 0.2)}$ & $\mathbf{[0.2, 0.4)}$ & $\mathbf{[0.4, 0.6)}$ & $\mathbf{[0.6, 0.8)}$ & $\mathbf{[0.8, 1.0)}$ & $\mathbf{=1.0}$          \\ \midrule
\textsc{Mistral-7B-v0.3}                 & 0.217                                       & 0.299                & 0.353        & 30.25\%                 & 30.25\%                  & 22.22\%                 & 8.23\%                 & 0.82\%                  & 8.23\%                           \\
\textsc{LLAMA-3.1-8B-Instruct}           & 0.559                                     & 0.638                & 0.618              & 7.21\%                 & 4.68\%                  & 29.09\%                 & 37.53\%                 & 8.83\%                 & 12.66\%                      \\
\textsc{LLAMA-3.1-70B-Instruct}          & 0.716                                        & 0.790                & 0.748            & 5.23\%                 & 2.10\%                  & 9.20\%                  & 32.17\%                  & 30.49\%                  & 20.81\%                     \\
\textsc{Qwen2.5-7B-Instruct}             & 0.771                                       & 0.828                & 0.820               & 2.12\%                 & 0.74\%                  & 5.33\%                 & 25.83\%                 & 38.66\%                 & 27.32\%                    \\
\textsc{Qwen2.5-32B-Instruct}            & 0.849                                      & 0.862                & 0.817          & 3.69\%                 & 0.95\%                  & 5.86\%                  & 19.74\%                 & 44.33\%                 & 25.44\%                          \\
\textsc{Qwen3-80B-Instruct}              & \underline{\textbf{0.925}}                 & \textbf{0.939}                 & \underline{0.925}        & 1.34\%                 & 0.05\%                  & 1.49\%                & 4.67\%                & 44.29\%                  & \underline{48.16\%}                               \\ \midrule
\textsc{Claude-Sonnet-4-6}               & 0.840                                      & 0.849                 & 0.846          & 2.07\%                 & 0.27\%                  & 3.37\%                 & 22.20\%                 & 41.02\%                 & 31.07\%                         \\
\textsc{GPT-4.1-mini}                    & 0.850                               & 0.862                & 0.832        & 3.23\%                  & 0.37\%                  & 4.55\%                  & 20.36\%                 & 43.52\%                 & 27.97\%                         \\ \midrule

{Human Learners}                           & 0.796                           & \underline{0.936}       & \textbf{0.942}                   & \textless{}0.01\%      & 0.68\%                  & 0.44\%                  & 3.15\%                  & 23.7\%                 & \textbf{72.7\%}    \\   
\specialrule{0.4pt}{2pt}{0pt}
\specialrule{1.0pt}{2pt}{0pt}
\end{tabular}
\caption{The accuracy and PSR results for all LLMs and human learners on all questions annotated by (XES$_{\rm full}$). For question-level PSR (Eq.~\ref{eq:psr_q}),  we report the proportion falling within different intervals. For aggregate PSR, we report both micro (Eq.~\ref{eq:psr_micro}) and macro (Eq.~\ref{eq:psr_macro}) results. For {Acc}, the proportion of questions with $\text{PSR} = 1.0$, and averaged PSR, we \textbf{bold} the best overall result and \underline{underline} the best result among LLMs.} \label{tab:psr}
\end{table*}

\paragraph{In-context scaffolding setup.}
For the scaffolding experiments (NB2), we provide each model with $k{=}3$ ICL exemplars selected under five strategies: 
(1) \emph{No context} (baseline); 
(2) \emph{Random}: three random questions of the same item type (FITB/MCQ); 
(3) \emph{Same-skill}: three questions sharing at least one concept with the target; 
(4) \emph{Similarity}: top-3 questions retrieved by BGE-M3 \cite{chen2024bge} embedding similarity; 
(5) \emph{Prerequisite}: three randomly sampled questions from the prerequisite set $\mathrm{Pre}(q)$. 
We use all questions in XES$_{\rm full}$ for this experiment.
For a fair comparison, all five conditions are evaluated on the {common subset} of questions where both prerequisite-based and concept-based selection yield $\geq 3$ eligible exemplars ($1{,}184$ FITB and $233$ MCQ questions). 
The prompt template is shown in Prompt \ref{prompt:icl_fitb}.

\section{Results}

\subsection{Prerequisite Satisfaction (NB1)}
\paragraph{Results on XES$_{\rm full}$.} We present the results in Table \ref{tab:psr}. 
The left part compares the accuracy and average PSR of LLMs and human learners. 
Overall, average PSR tends to increase with accuracy, which is expected because PSR approximates the conditional probability of correctly answering prerequisite questions given that their advanced question is solved, and is therefore positively correlated with overall correctness probability. 
Nevertheless, human learners achieve a high PSR of $0.936/0.942$ at a relatively modest accuracy of $0.796$. 
In contrast, \textsc{Qwen2.5-32B-Instruct}, despite outperforming human learners in accuracy, obtains substantially lower PSR scores. Only \textsc{Qwen3-80B-Instruct}, with a notably higher accuracy of $0.925$, slightly surpasses human learners in micro-PSR but remaining lower in macro-PSR. These results suggest that even when LLMs match or exceed human accuracy, their response patterns generally remain less consistent with the prerequisite structure.

The right portion of Table~\ref{tab:psr} presents a more fine-grained view through the distribution of question-level PSR across intervals. 
We separately highlight the proportion of questions with $\text{PSR}=1.0$, which represents perfectly coherent knowledge for that question that strictly satisfies the closure property of KST. 
This distributional analysis reveals a deeper gap. 
For human learners, the vast majority of questions (72.7\%) achieve a perfect PSR of 1.0, indicating that human correct answers are almost always grounded in mastery of the prerequisite knowledge. 
As model capability increases, the LLM distributions shift toward the higher PSR intervals; nevertheless, perfect-PSR rates remain substantially below that of human learners. Even the best-performing model, \textsc{Qwen3-80B-Instruct}, reaches 48.16\%, exhibiting less consistent full prerequisite satisfaction.
\begin{table}[]
\small
\centering
\setlength{\tabcolsep}{3pt}
\begin{tabular}{@{}lcccH@{}}
\specialrule{1.0pt}{0pt}{2pt}
\specialrule{0.4pt}{0pt}{2pt}
\multirow{2}{*}{\textbf{Model}} & \multicolumn{1}{c}{\multirow{2}{*}{\textbf{Acc.}}} & \multicolumn{3}{c}{\textbf{PSR}}                                                                                 \\
                                & \multicolumn{1}{c}{}                                   & \multicolumn{1}{c}{\textbf{Micro}} & \multicolumn{1}{c}{\textbf{Macro}} & \multicolumn{1}{c}{\textbf{=1.0}} \\ \midrule
\textsc{Mistral-7B-v0.3}             & 0.217                                                  & 0.325                              & 0.387                              & 0\%                                      \\
\textsc{LLAMA-3.1-8B-Inst.}           & 0.560                                                  & 0.616                              & 0.614                              & 5.63\%                                  \\
\textsc{LLAMA-3.1-70B-Inst.}          & 0.717                                                  & 0.826                              & 0.789                              & 32.1\%                                 \\
\textsc{Qwen2.5-7B-Inst.}            & 0.771                                                  & 0.822                              & 0.834                              & 21.3\%                                \\
\textsc{Qwen2.5-32B-Inst.}           & 0.849                                                  & 0.848                              & 0.802                              & 22.1\%                                   \\
\textsc{Qwen3-80B-Inst.}         & \textbf{\underline{0.925}}                                                  & \underline{0.946}                              & \underline{0.932}                              & \underline{55.7}\%                                   \\ \midrule
\textsc{Claude-sonnet-4-6}               & 0.841                                                  & 0.863                              & 0.858                              & 28.1\%                                   \\
\textsc{GPT-4.1-mini}                    & 0.849                                                  & 0.852                              & 0.822                              & 21.5\%                                  \\ \midrule
Human Learners                          & 0.796                                                  & \textbf{0.950}                     & \textbf{0.951}                     & \textbf{81.6\%}                          \\ 
\specialrule{0.4pt}{0pt}{2pt}
\specialrule{1.0pt}{0pt}{2pt}
\end{tabular}
\caption{Accuracy and PSR results on the verified subset XES$_{\rm verified}$. We \textbf{bold} the best overall result and \underline{underline} the best result among LLMs.}\label{tab:psr_verified}
\end{table}

\paragraph{Results on XES$_{\rm verified}$} 
To assess how LLM annotation errors affect our results, we repeat the evaluation on the verified subset (Table \ref{tab:psr_verified}). The main conclusions hold: micro-PSR and macro-PSR follow the same trends as on the full LLM-annotated set, and the proportion of questions with PSR = 1 shows the same overall pattern, with human learners at 81.6\% against 55.7\% for the best-performing LLM.
Both groups achieve higher PSR = 1 rates than on the full set, as expected because the verified subset has fewer prerequisites per target question (7.69 on average, vs.\ 16.43), making the all-prerequisites-correct condition easier to satisfy. Nevertheless, the human--LLM gap remains comparable (25.9 vs.\ 24.54 percentage points), indicating that this change in evaluation scale does not alter the underlying result.

Taken together, the results on both XES$_{\rm full}$ and XES$_{\rm verified}$ datasets demonstrate the same conclusion that human knowledge structures are more coherent than those of current LLMs.
In addition, the consistent findings indicate that PSR is robust to a moderate level of concept annotation error.

\researchquestionbox{\textbf{Finding on NB1}: Despite achieving comparable or higher accuracy than humans, LLMs exhibit substantially lower prerequisite satisfaction, with far fewer correct answers being fully grounded in the requisite knowledge structure.}{blue}

\subsection{Scaffolding Effect (NB2)}
To evaluate NB2, we compare five retrieval strategies for selecting in-context exemplars: (1) no-context baseline, (2) random questions, (3) questions sharing the same skill, (4) prerequisite questions identified by our knowledge graph, and (5) semantically similar questions. All methods are evaluated on the common subset where both prerequisite and same-skill exemplars are available. We report task accuracy and Scaffolding Gain (SG), defined in Eq.~\ref{eq6} as the change in accuracy relative to the no-context baseline.

In human learning, prerequisite or related knowledge facilitates the acquisition and application of more advanced concepts~\citep{wood1976role}, and revisiting foundational concepts often improves performance on dependent tasks~\citep{sweller1988cognitive, falmagne2013knowledge}. 
If LLMs organize math knowledge in similar prerequisite dependencies, providing prerequisite examples should therefore yield larger gains than others. 

\begin{figure}[t]
\centering
\includegraphics[width=\linewidth]{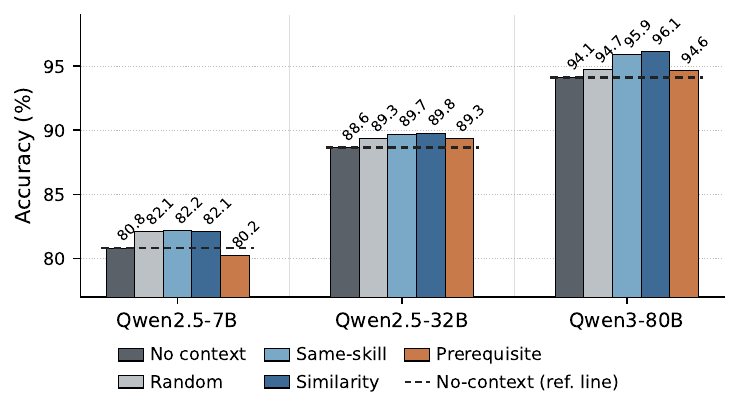}
\caption{
Knowledge scaffolding across various context types on the common subset.} \label{fig:nb3}
\end{figure}

Figure~\ref{fig:nb3} shows that providing additional context generally improves LLM performance. However, prerequisite-based context does not consistently outperform alternative retrieval strategies. 
While most in-context strategies improve performance relative to the no-context baseline, the strongest gains consistently come from same-skill and semantically similar exemplars. Even randomly selected examples frequently match the effectiveness of prerequisite retrieval. 
For \textsc{Qwen2.5-7B-Instruct}, prerequisite contexts provide no measurable benefit and slightly reduces performance relative to the no-context baseline (SG=$-0.56$). 
For \textsc{Qwen2.5-32B-Instruct} and \textsc{Qwen3-80B-Instruct}, prerequisite retrieval yields only modest gains (${\rm SG}=+0.70$ and $+0.50$, respectively), remaining below same-skill and semantically similar retrieval; further, in \textsc{Qwen2.5-32B-Instruct}, it merely matches random exemplars.

These findings suggest that the benefits of in-context examples arise primarily from exposure to relevant solution patterns rather than from activating prerequisite knowledge required by the target problem. 
While prerequisite examples can sometimes improve performance, they do not provide a systematic advantage over same-skill or semantically similar contexts. 
This contrasts with the central prediction of KST, where prerequisite knowledge plays a privileged role in supporting downstream learning. Overall, the results suggest that current LLMs rely more on contextual pattern matching than on a structured prerequisite hierarchy during mathematical reasoning.

\researchquestionbox{\textbf{Finding on NB2}: Contrary to the premise of KST, contexts of relevant knowledge do not provide stronger scaffolding than semantically-related examples, suggesting that LLMs benefit primarily from contextual pattern matching rather than from activating relevant knowledge during reasoning.}{orange}

\subsection{Knowledge Subsumption (NB3)}
In this experiment, we investigate whether the knowledge state of a stronger model subsumes that of a weaker model, as prescribed by NB3. 
Similar to previous experiments, we include human learners as a reference. 
To compare the knowledge states of different learners, we simulate three human groups of \textit{low}, \textit{medium}, and \textit{high} ability.
Specifically, we first compute the average accuracy of each student across all attempted questions, and partition students into three groups based on performance percentiles: the bottom 0--30\% as the \textit{low} group, 30--60\% as the \textit{medium} group, and 60--90\% as the \textit{high} group. 
The accuracy scores for three groups are $0.64$, $0.80$, and $0.89$, respectively.
We exclude the top 10\% of students, as their near-perfect knowledge states would trivially yield close to 100\% overlap.  The knowledge state of each human group is then defined as the set of questions whose average accuracy within that group exceeds a threshold of $0.5$.

\begin{figure}[t]
\centering
\includegraphics[width=1\columnwidth]{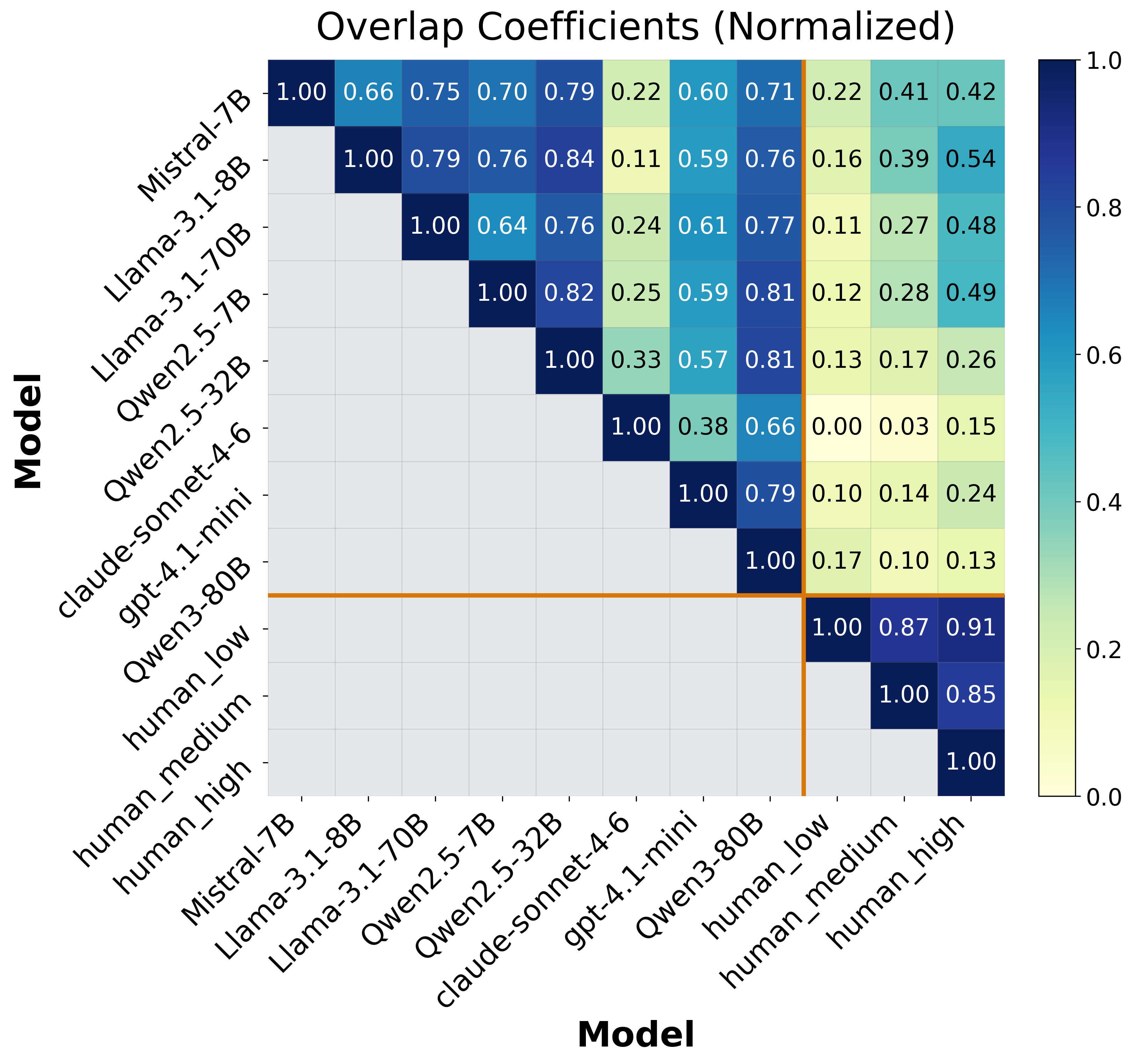}
\caption{Normalized knowledge overlap coefficients (Eq. \ref{eq:koc_norm}) between the knowledge states of different LLMs and of different human learner groups. Models are ordered by performance from weaker to stronger within both the LLM and human groups.} \label{fig:overlap}
\label{fig:overlap_coefficients}
\end{figure}

We present the normalized ${\rm KOC}_{\rm norm}$ results across models and human groups in Figure~\ref{fig:overlap}, and the unnormalized results can be found in Appendix Figure \ref{fig:app_koc_raw}.  
From results across different regions (by orange lines), we can observe that human learner groups of different performance levels (lower-right) exhibit high overlap, consistent with our expectation that in a structured and interdependent knowledge system, the knowledge of weaker learners should be largely subsumed by that of stronger ones. An interesting finding is that the overlap between LLMs and human learners (upper-right) is substantially lower, and stronger models appear to show even less alignment with human learners.

Among LLMs (upper-left region), the normalized KOC reveals more nuanced patterns. 
Open-source models exhibit relatively strong mutual overlap, though still lower than that among human learner groups. 
In contrast, the two closed-source models, \textsc{GPT-4.1-mini} 
and \textsc{Claude}, show considerably lower overlap with open-source models, and only 0.38 overlap with each other. 
A plausible reason is that open-source models share substantial portions of their training corpora, while closed-source models likely draw from more diverse and proprietary data sources, leading to divergent knowledge distributions.

Taken together, these results suggest that LLMs do not necessarily follow a consistent shared knowledge structure. 
The locally high overlap observed among some model pairs is more likely a reflection of shared 
training data than evidence of a coherent common knowledge organization.

\researchquestionbox{\textbf{Finding on NB3}: Human learners exhibit near-perfect knowledge subsumption across ability levels, while LLMs show substantially lower and more fragmented knowledge growth — with stronger LLMs paradoxically diverging further from human knowledge progression patterns.}{green}

\section{Conclusion}

We introduced a KST–grounded framework that evaluates the structural coherence of LLM knowledge via various normative behaviors. 
Across eight LLMs and more than 18,000 human learners, we find that high accuracy masks pervasive structural inconsistency: even the strongest model fully satisfies prerequisites for only $48.16\%$ of its correct answers, compared to $72.7\%$ for human learners. Moreover, providing prerequisite-grounded context yields no clear advantage over surface-similar baselines, indicating that LLMs do not reliably use prerequisite knowledge as human-like scaffolding. These findings suggest that current LLM knowledge is fragmented rather than hierarchical, and motivate structure-aware assessment as a complementary lens for rigorous evaluation.

\section{Limitations}
We state the limitations of this work from the following aspects.
First, our framework assumes the availability of an expert-defined concept dependency graph. While such resources exist for mathematics and several educational domains, constructing reliable prerequisite structures may be challenging in domains where knowledge dependencies are less explicit or less well documented. Second, we focus exclusively on mathematics, a domain with relatively well-established prerequisite relations. Whether the same observations extend to other domains, such as science, programming, or general factual knowledge, remains an open question.
Finally, we operationalize knowledge states at the question level rather than directly modeling latent concept mastery. Although this enables large-scale evaluation using observable responses, question-level correctness is only an imperfect proxy for underlying knowledge states. Future work could incorporate concept-level mastery estimation to more closely align the evaluation with the original formulation of Knowledge Space Theory.

\section{Ethical Statement}

This work investigates the knowledge structure of large language models using publicly available mathematics datasets and benchmark questions. All data are used only for research purposes and do not contain personal or sensitive information. AI assistance was employed for language editing and proofreading.

\section*{Acknowledgments}
This research was supported by the Swiss National Science Foundation (SNSF) under grant number 10009282 and by a Swiss AI large grant. Heejin Do was also supported by the ETH AI Center through an ETH AI Center postdoctoral fellowship to H.

\bibliography{custom}

\appendix
\newpage
\section{Prompts}\label{Appendix:prompts}

\begin{promptbox}[label={prompt:baseline_fitb}]{LLM Evaluation Prompt}

\textbf{System:} \\
Please reason step by step, and put your final answer within \texttt{\textbackslash boxed\{\}}. \\

\textbf{User:} \\
\{Question\}
\end{promptbox}

\begin{promptbox}[label={prompt:skill_extraction}]{Concept Extraction Prompt}

You are an expert in mathematics education. You will be given a list of reference skills and a mathematical question along with its solution. \\
Your task is to identify all the skill(s) from the given skill list that are required to solve the problem. \\

Do NOT invent new skills. \\
Do NOT modify skill names. \\
Only choose from the provided skill list. \\

Output a \texttt{JSON} array containing only the ids of selected skills. If no skill from the list applies, output an empty list: [] \\

Output ONLY valid \texttt{JSON} without any additional text. \\

\textbf{Skill list}: \{Reference skills\}

\textbf{Question:} \{Question\}

\textbf{Solution}: \{Solution\}

\end{promptbox}

\begin{promptbox}[label={prompt:icl_fitb}]{Scaffolding In-Context Prompt}

\textbf{System:} \\
Please reason step by step, and put your final answer within \texttt{\textbackslash boxed\{\}}. \\

\textbf{User:} \\
The following problems and their solutions are provided as background knowledge: \\

\textbf{Problem 1:} \\
\{Question 1\} \\
\textbf{Options:} \{Options 1\} \textcolor{red!70}{\% MCQ-type Only} \\
\textbf{Solution:} \{Solution 1\} \\
\textbf{Answer:} \{Answer 1\} \\

\textbf{Problem 2:} \\
\{Question 2\} \\
\textbf{Options:} \{Options 2\} \textcolor{red!70}{\% MCQ-type Only} \\
\textbf{Solution:} \{Solution 2\} \\
\textbf{Answer:} \{Answer 2\} \\

\textbf{Problem 3:} \\
\{Question 3\} \\
\textbf{Options:} \{Options 3\} \textcolor{red!70}{\% MCQ-type Only} \\
\textbf{Solution:} \{Solution 3\} \\
\textbf{Answer:} \{Answer 3\} \\

\texttt{---} \\

Now solve the main problem: \\
\{Target Question\} \\

\textcolor{red!70}{\% MCQ-type Only} \\
\textbf{Answer Choices:} (A) \{Target Option A\} (B) \{Target Option B\} (C) \{Target Option C\} (D) \{Target Option D\} 
\end{promptbox}

\begin{table*}[]
\small
\centering
\begin{tabular}{p{0.2cm}p{7cm}p{7.5cm}}
\specialrule{1.0pt}{0pt}{2pt}
\specialrule{0.4pt}{0pt}{2pt}
\textbf{ID} & \textbf{Question}                                                                                                                                                                                             & \textbf{LLM-annotated concepts}                                                                                                                                                                                                                                                                                                                                                          \\ \midrule
1  & Distance between A and B is 350 km. A car leaves from A at 8:00 with speed 40 km/h toward B. After 2 hours, another car leaves from B toward A at 50 km/h. At what time do the two cars meet on the road? ( ) & Solve two-step word problems using the four operations. Represent these problems using equations with a letter standing for the unknown quantity. Assess the reasonableness of answers using mental computation and estimation strategies including rounding.                                                                                                                            \\
2  & A rectangle with perimeter \$\$20\$\$ meters and length \$\$7\$\$ meters; what is its width ( ) meters?                                                                                                               & Solve real world and mathematical problems involving perimeters of polygons, including finding the perimeter given the side lengths, finding an unknown side length, and exhibiting rectangles with the same perimeter and different areas or with the same area and different perimeters.                                                                                               \\
3  & From the numbers \$\$5\$\$, \$\$6\$\$, and \$\$7\$\$, choose two numbers to form the largest two-digit number ( ).                                                                                                        & Understand that the two digits of a two-digit number represent amounts of tens and ones.                                                                                                                                                                                                                                                                                                 \\ \midrule
4  & There are \$\$8\$\$ boxes of apples, with \$\$40\$\$ apples per box. If evenly distributed among \$\$4\$\$ classes, how many per class?                                                                                   & Use multiplication and division within 100 to solve word problems involving equal groups, arrays, and measurement quantities, for example by using drawings and equations with a symbol representing the unknown number.                                                                                                                                                                 \\
5  & A fraction with denominator 6 and numerator 5 is \_\_\_\_.                                                                                                                                                    & Understand a fraction 1/b as the quantity formed by one part when a whole is divided into b equal parts, and understand a fraction a/b as the quantity formed by a parts of size 1/b.                                                                                                                                                                                                    \\
6  & On one side of a road that is \$\$24\$\$ meters long, plant a tree every \$\$4\$\$ meters, with neither end planted. How many trees can be planted along this road?                                                   & Express the length of an object as a whole number of length units by laying multiple copies of a shorter object (the length unit) end to end. Understand that the length measurement of an object is the number of same-size length units that span it with no gaps or overlaps. Limit to cases where the object is measured by a whole number of length units with no gaps or overlaps. \\ 
\specialrule{0.4pt}{2pt}{0pt}
\specialrule{1.0pt}{2pt}{0pt}
\end{tabular}
\caption{Examples of questions and their LLM-annotated concepts on the XES dataset.}\label{tab:case}
\end{table*}

\begin{figure}[t]
\centering
\includegraphics[width=1\columnwidth]{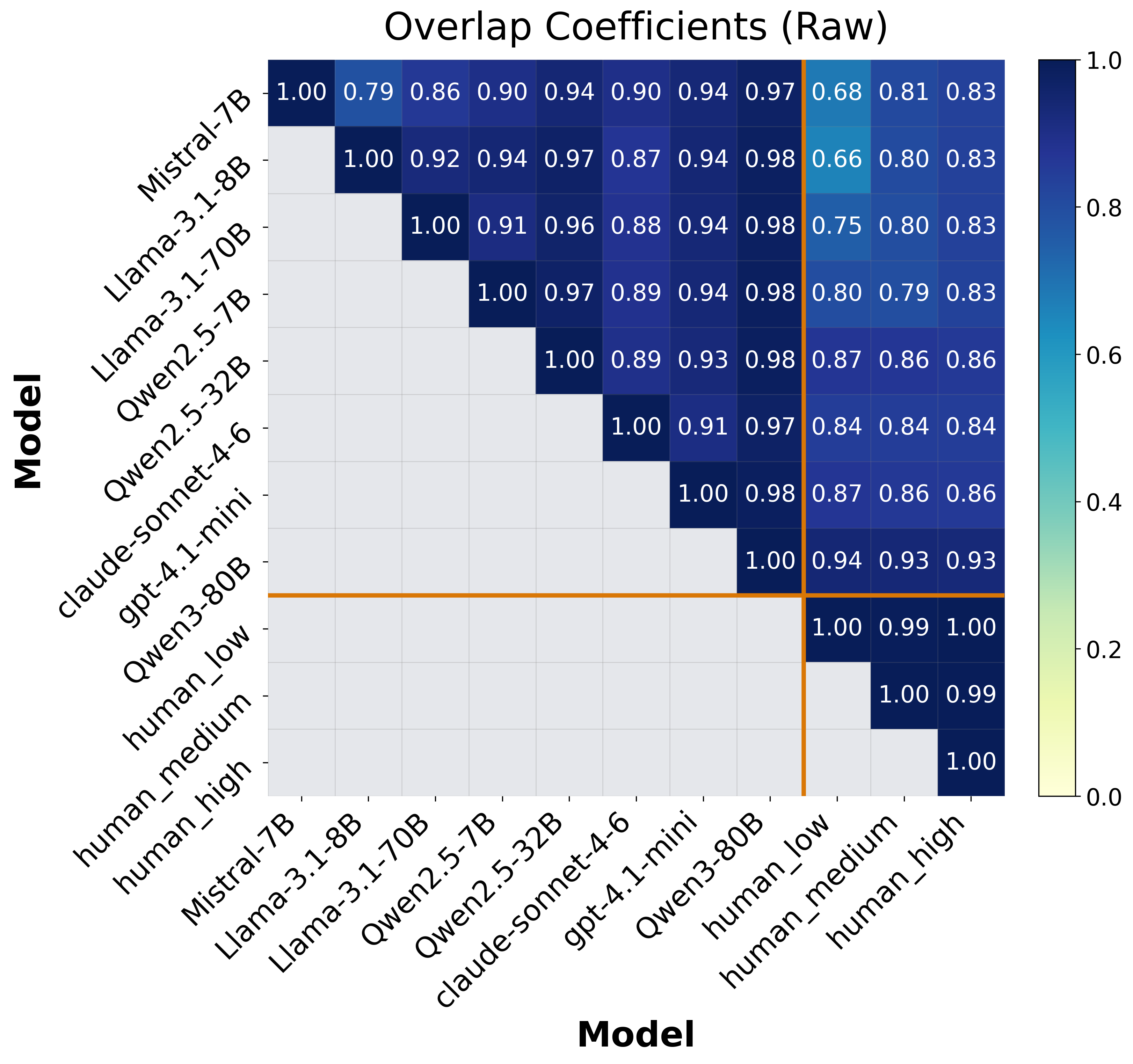}
\caption{Raw knowledge overlap coefficients (Eq. \ref{eq:koc}) between the knowledge states of different LLMs and human learner groups.} \label{fig:app_koc_raw}
\end{figure}

\section{Evaluation of LLM-based Concept Annotation on NYS Example Questions} \label{app:annotation_eval}

\paragraph{Quantitative Evaluation on NYS Standards.}
Among the 480 NYS mathematics concept descriptions, 325 are associated with an example problem. 
We treat these example {<Problem, Concept>} pairs as ground-truth annotations and evaluate the accuracy of LLM-based concept annotation on them.
Out of the 325 instances, 192 are correctly annotated, 74 are incorrectly annotated, and the remaining 59 are left unannotated, i.e., no matching concept was identified. This suggests that the LLM-based annotation has some limitations in recall.
However, in our framework, questions without an identified concept are discarded and excluded from subsequent computations. 
As a result, this may reduce the number of prerequisite relations we are able to discover, but we prioritize the precision of discovered prerequisite relations over introducing noisy ones.
Lower recall also implies that the true proportion of ${\rm PSR}=1$ cases in Table \ref{tab:psr} is likely overestimated for both humans and LLMs. 
Excluding cases without identified concepts, the annotation accuracy of the LLM reaches 72\%. There remains substantial room for improvement, which we expect could be achieved with stronger annotation LLMs.

\paragraph{Case Study on XES.} We list several examples of both high-quality and imperfect concept annotations on our XES dataset in Table \ref{tab:case}. Examples 1–3 illustrate accurate annotations with strong item–concept alignment. The remaining examples correspond to cases that are still related to the items, but exhibit imperfect alignment in different ways. In Example 4, the annotated concept is conceptually relevant, but involves numerical values beyond 100, exceeding the scope of the item itself. Example 5 is associated with a concept that is substantially broader than the competency required by the question. In Example 6, the concept captures the underlying idea of equal-length partitioning, but the item itself primarily requires discrete interval counting with endpoint constraints rather than length measurement.

\section{Reasoning Scores and PSR Provide Complementary Views of Capability}
\label{app:reasoning-saturation}

A natural question is whether existing reasoning-quality metrics capture the same information as prerequisite satisfaction. To investigate, we score each model's reasoning traces using \textsc{GPT-4.1-mini} as an LLM-as-judge along three dimensions: \textit{Relevance}, \textit{Coherence}, and \textit{Accuracy}, as defined by \citet{do2025defines}.

\begin{figure}[h]
\centering
\includegraphics[width=\linewidth]{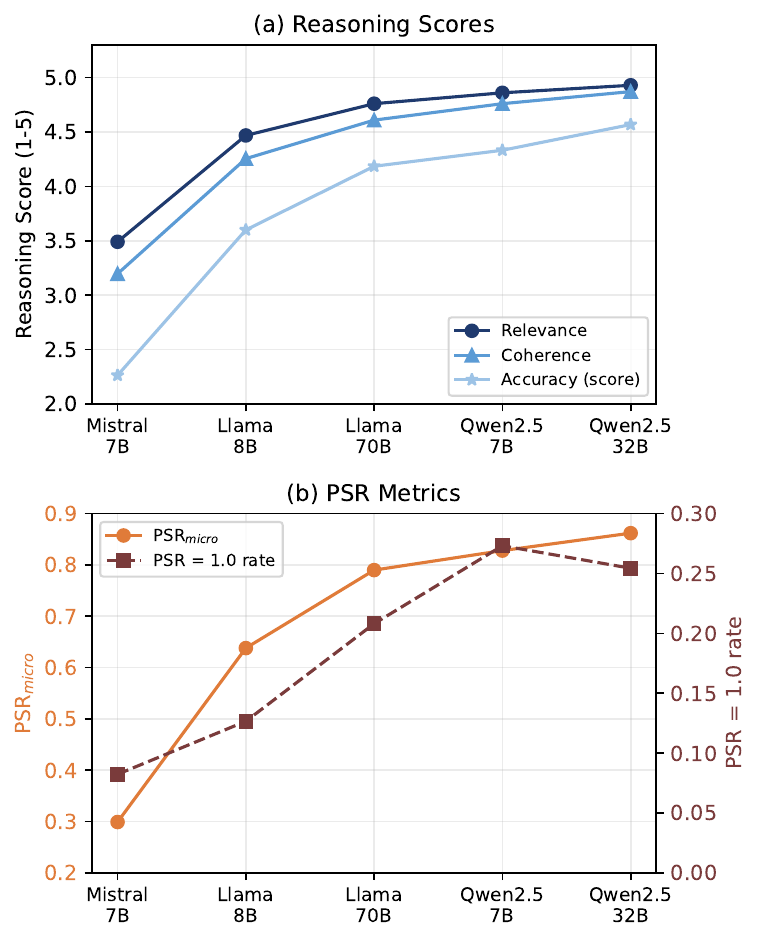}
\caption{
Comparison of reasoning scores (a) and prerequisite satisfaction (b) across five open-source models. 
} \label{fig:reasoning}
\end{figure}

Figure~\ref{fig:reasoning} compares reasoning scores and PSR across the five open-source models. Although both metric families generally improve with model capability, they exhibit different patterns in the high-performance regime. Among the three strongest models, reasoning scores differ only modestly, whereas the PSR$=1.0$ rate remains more variable and non-monotonic. Specifically, Relevance increases from 4.76 for \textsc{Llama-3.1-70B-Instruct} to 4.86 for \textsc{Qwen2.5-7B-Instruct} and 4.93 for \textsc{Qwen2.5-32B-Instruct}, while the corresponding PSR$=1.0$ rates are 20.81\%, 27.32\%, and 25.44\%, respectively.

This divergence reflects the different behaviors captured by the two metrics. LLM-as-judge evaluation measures the local relevance, coherence, and correctness of individual reasoning traces, whereas PSR measures cross-question consistency with prerequisite relations. Consequently, models that appear similarly strong under conventional reasoning evaluation may still differ in strict prerequisite satisfaction. Therefore, we view PSR as a complementary diagnostic: reasoning scores capture the quality of individual reasoning traces, while PSR captures whether a model's response patterns consistently respect the prerequisite structure.

\end{document}